\documentclass[letterpaper]{article} % DO NOT CHANGE THIS
\usepackage{aaai24}  % DO NOT CHANGE THIS
\usepackage{times}  % DO NOT CHANGE THIS
\usepackage{helvet}  % DO NOT CHANGE THIS
\usepackage{courier}  % DO NOT CHANGE THIS
\usepackage{xcolor}
\usepackage[utf8]{inputenc}
\usepackage{tikz}
\usepackage{pgfplots}
\usepackage{subcaption}
\usepackage{pgfplots}
\usepackage{algorithm}
\usepackage{multirow}
\usepackage{booktabs}
\usepackage{array}
\usepackage{algorithmic}
\usepackage{amsmath} % for \min, \max, \text
\usepackage[hyphens]{url}  % DO NOT CHANGE THIS
\usepackage{graphicx} % DO NOT CHANGE THIS
\usepackage{natbib}  % DO NOT CHANGE THIS AND DO NOT ADD ANY OPTIONS TO IT
\usepackage{caption} % DO NOT CHANGE THIS AND DO NOT ADD ANY OPTIONS TO IT
\usepackage{algorithm}
\usepackage{algorithmic}
\usepackage{newfloat}
\usepackage{listings}
\DeclareCaptionStyle{ruled}{labelfont=normalfont,labelsep=colon,strut=off} % DO NOT CHANGE THIS
\floatstyle{ruled}
\newfloat{listing}{tb}{lst}{}
\floatname{listing}{Listing}
\title{Suppressing Uncertainty in Gaze Estimation}
\author {
    Shijing Wang,
    Yaping Huang\thanks{Corresponding author}
}
\affiliations {
    Beijing Key Laboratory of Traffic Data Analysis and Mining, Beijing Jiaotong University, China\\
    \{shijingwang, yphuang\}@bjtu.edu.cn
}

\usepackage{bibentry}
\begin{document}
\maketitle

\begin{abstract}
Uncertainty in gaze estimation manifests in two aspects: 1) low-quality images caused by occlusion, blurriness, inconsistent eye movements, or even non-face images; 2) incorrect labels resulting from the misalignment between the labeled and actual gaze points during the annotation process. Allowing these uncertainties to participate in training hinders the improvement of gaze estimation.  To tackle these challenges, in this paper, we propose an effective solution, named Suppressing Uncertainty in Gaze Estimation (SUGE), which introduces a novel triplet-label consistency measurement to estimate and reduce the uncertainties. Specifically, for each training sample, we propose to estimate a novel ``neighboring label'' calculated by a linearly weighted projection from the neighbors to capture the similarity relationship between image features and their corresponding labels, which can be incorporated with the predicted pseudo label and ground-truth label for uncertainty estimation. By modeling such triplet-label consistency, we can measure the qualities of both images and labels, and further largely reduce the negative effects of unqualified images and wrong labels through our designed sample weighting and label correction strategies. Experimental results on the gaze estimation benchmarks indicate that our proposed SUGE achieves state-of-the-art performance.
\end{abstract}

\section{Introduction}
Gaze estimation is a crucial task in computer vision that aims to accurately determine the direction of a person's gaze based on visual cues. In recent years, gaze estimation has gained significant attention due to its wide-ranging applications in fields such as human-computer interaction~\cite{majaranta2014eye}~\cite{rahal2019understanding}, virtual reality~\cite{patney2016perceptually}~\cite{kim2019nvgaze}, and assistive technology~\cite{jiang2017learning}~\cite{liu2016identifying}~\cite{dias2020gaze}. Benefiting from the deep learning techniques and large-scale training data, appearance-based gaze estimation has made rapid progress and achieved promising results.

\begin{figure}
\centering
\includegraphics[width=8.0cm]{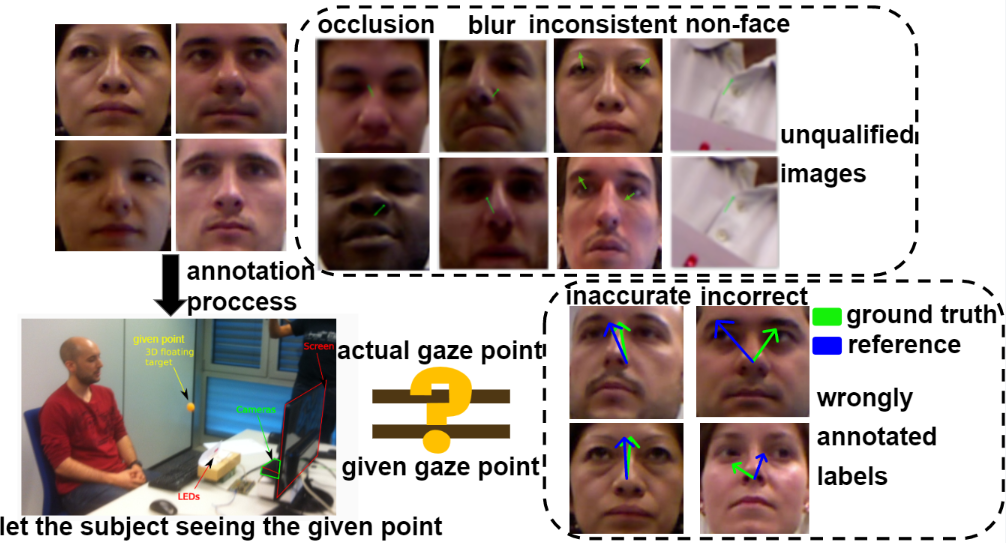}
\caption{Illustration of uncertainties in gaze estimation using the EyeDiap dataset as an example. The upper half of the figure reflects unqualified images, where the images in right side are extremely difficult for machines and even human. These images are better to be suppressed in training. The lower half of the figure reflects wrongly annotated labels, where the left side represents the common data annotation process. Due to the challenge of achieving perfect alignment between the actual gaze points and the given gaze points, inaccurate and incorrect labels exist in the datasets, which should be rectified.}
\label{Fig1.motivation}
\end{figure}

\renewcommand{\floatpagefraction}{.9}

Recent works are mainly dedicated to developing advanced networks~\cite{zhang2017s}~\cite{kellnhofer2019gaze360}~\cite{cheng2020coarse}~\cite{cheng2022gaze} to extract distinctive gaze features. However, the success of deep models heavily relies on sufficient amount of data with accurate ground truth labels. Unfortunately, it is very challenging for humans to provide consistent and precise annotations for gaze estimation task, especially in the complex natural scenes. 
Common gaze estimation datasets~\cite{funes2014eyediap}~\cite{zhang2017mpiigaze}~\cite{kellnhofer2019gaze360}~\cite{zhang2020eth} usually obtain annotations by requiring subjects to fixate on given points during data collection. But this annotation strategy is based on the assumption of the perfect alignment between the actual gaze points and the provided points, which is practically impossible to achieve.

As shown in Fig.~\ref{Fig1.motivation}, many captured gaze images suffer from serious quality degradation due to eyelid occlusion, blurriness of the eyes, inconsistent eye movements, which may lead to some unqualified eye images, or even background images are involved in the collecting process. On the other hand, there are also a significant number of visibly incorrect labels presented in the commonly used 
datasets. Allowing these unqualified images and incorrect ground truth labels to be included in the training process may result in overfitting, which hinders the model from learning the discriminative features for accurate gaze estimation. 
Generally, although many efforts have been made for getting precise annotations, the noisy data and labels are inevitably introduced, which is neglected in the previous works. 

\begin{figure}
\centering
\includegraphics[width=8cm]{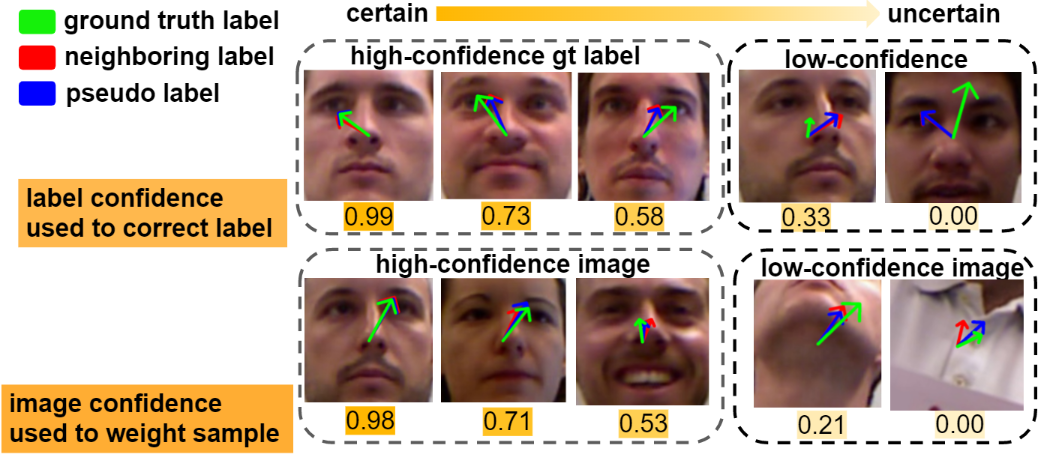}
\caption{Visualization results of varying image and label confidences using samples from folds 0, 1, 3 of the EyeDiap dataset as the training set in the subsequent epoch after the warm-up phase. These results showcase the effectiveness of the two uncertainty metrics we design.}
\label{Fig3.confidence effectiveness}
\end{figure}

To address the above-mentioned issues, in this paper, we propose a novel solution, named ``Suppressing Uncertainty in Gaze Estimation'' (SUGE), which brings a new perspective of uncertainty estimation for gaze estimation task. The key issue of our work is how to effectively estimate the uncertainty for measuring the quality of images and labels, and meanwhile alleviating the negative effects of them. To achieve this goal, we propose a novel triplet-label consistency method to estimate the uncertainty, where ``neighboring label'' is proposed by computing a weighted average of labels from neighboring image features, and coupled with the predicted labels and ground truth labels to 
calculate two uncertainty metrics of each training sample by using Gaussian Mixture Model (GMM). As shown in Fig.~\ref{Fig3.confidence effectiveness}, by modeling two uncertainty metrics, we can obtain two confidences: image confidence and label confidence, where the former reflects the quality of the images and the latter measures the correctness of the annotations. Afterwards, we utilize the estimated confidences for guiding the further training process: the image confidence is used to weight the training sample, and the label confidence is referred when performing label correction with the predicted pseudo-label and neighboring label. The effectiveness of our proposed SUGE is comprehensively evaluated on popular gaze estimation benchmarks.

In summary, our contribution is three-fold:
\begin{itemize}
    \item We address the gaze estimation task from a new perspective of uncertainty estimation to mitigate the effect of low-quality images and incorrect annotations, which commonly exists in real-world gaze estimation datasets but is ignored in previous works.
    \item We propose a novel triplet-label consistency measurement to estimate the uncertainty, where a novel ``neighboring label'' representing the local consistency is coupled with the predicted pseudo label and ground truth label to assess the uncertainty of samples. Then the produced uncertainty is further utilized for better training by the proposed label correction and sample weighting strategies.
    \item We conduct comprehensive experiments on real-world gaze estimation benchmarks, and the experimental results demonstrate that our method achieves state-of-the-art performance.
\end{itemize}

\section{Related Works}

\subsection{Gaze Estimation}

% Methods for gaze estimation can be roughly divided into two main types~\cite{hansen2009eye}: model-based and appearance-based. Model-based methods~\cite{zhu2007novel}~\cite{valenti2011combining}~\cite{alberto2014geometric} rely on precise 3D models of the eye, making them sensitive to controlled experimental settings and necessitating sophisticated data acquisition equipment. In contrast, appearance-based methods~\cite{cheng2021appearance} adopt machine learning techniques to directly map the images captured by cameras to gaze directions. With the development of deep learning, appearance-based methods have gained substantial attention due to their generality and ease of implementation in recent years.

Gaze estimation methods are categorized into model-based and appearance-based approaches~\cite{hansen2009eye}. Model-based techniques~\cite{zhu2007novel,valenti2011combining,alberto2014geometric} utilize 3D eye models with specialized equipment in controlled settings, while appearance-based methods~\cite{cheng2021appearance} employ machine learning to map images to gaze directions, gaining popularity for their adaptability.

% Existing appearance-based methods primarily concentrate on refining the model architecture under different input scenarios. Earlier studies~\cite{zhang2015appearance}~\cite{cheng2018appearance} utilize eye images and given head poses as inputs, successfully integrating deep learning in gaze estimation research. Subsequently, to address potential errors in given head poses, some studies~\cite{krafka2016eye}~\cite{zhu2017monocular} use both eye and head images as inputs to achieve joint prediction of gaze and head poses. More recently, some works~\cite{kellnhofer2019gaze360}~\cite{zhang2020eth}~\cite{cheng2022gaze} directly use head images as input and leverage the powerful learning capabilities of advanced general-purpose deep models, achieving state-of-the-art performance without the need for additional design.
Data annotation poses a critical challenge for appearance-based gaze estimation. Two primary methods are prevalent. The first involves model-based eye-tracking devices, such as desktop eye trackers~\cite{park2020towards} or head-mounted ones~\cite{fischer2018rt}, while the former is restricted to specific distances and head poses, and the latter severely occludes eye appearance. The second method employs fixation-based annotation scheme, allowing subjects to focus on specific points of interest~\cite{CAVE_0324,sugano2014learning,funes2014eyediap,zhang2017mpiigaze,kellnhofer2019gaze360,zhang2020eth}. This approach remains flexible, unaffected by appearance changes, making it cost-effective and widely adopted for creating gaze estimation datasets.
% Apart from designing advanced architectures, another critical challenge for gaze estimation lies in accurately annotating the data. Upon investigating numerous gaze estimation datasets, we identifies two primary annotation methodologies. The first approach involves the utilization of model-based eye-tracking devices for data collection. These devices can be categorized into two types based on their wearability. Desktop eye trackers~\cite{park2020towards} are limited by a restricted working distance and a constrained range of head poses. Conversely, head-mounted eye trackers~\cite{fischer2018rt} offer unrestricted head movement, but they may severely alter the appearance of the eyes due to the wearable nature of the device. The second approach employs fixation-based annotation, where subjects are instructed to focus their gaze on specific points of interest. This fixation-based methodology~\cite{CAVE_0324}~\cite{sugano2014learning}~\cite{funes2014eyediap}~\cite{zhang2017mpiigaze}~\cite{kellnhofer2019gaze360}~\cite{zhang2020eth} allows for more flexibility, as it is not constrained by environmental factors or head poses, and does not introduce any changes to the eye's appearance. Furthermore, this method is cost-effective, as it does not require sophisticated eye-tracking equipment, making it a popular choice for creating gaze estimation datasets.

Despite great efforts have been made in building gaze estimation datasets, the complexity of generating gaze annotations inevitably introduces low-quality data and labels. These issues are often overlooked in existing works and have become significant bottlenecks hindering the development of gaze estimation algorithms. Our work represents the first attempt to address the problem of annotation quality from a novel perspective of uncertainty estimation.

\subsection{Uncertainty Estimation in Computer Vision}

Uncertainty estimation is crucial for capturing the randomness in the learning process across computer vision tasks, such as facial expression recognition~\cite{wang2020suppressing}, saliency detection~\cite{zhang2020uc}, and edge detection~\cite{zhou2023treasure}. For gaze estimation, handling data uncertainty is crucial. Approaches like~\cite{kellnhofer2019gaze360,dias2020gaze} incorporate uncertainty prediction heads at the output layer to handle high uncertainty data. Nonaka et al.~\cite{nonaka2022dynamic} focus on multi-source input uncertainty by adding prediction heads for each input's features. Recent work~\cite{cai2023source} tackles cross-domain adaptation by enhancing image quality to mitigate image uncertainty and reducing prediction variance to manage model uncertainty.

% In contrast, our study reveals the prevailing challenges arising from low-quality samples and inaccurately annotated labels simultaneously in the current gaze estimation datasets, which has not been thoroughly explored previously. To address these issues, we propose a novel approach that assesses the consistency among three types of labels to estimate the underlying uncertainty in image and label space, allowing us to mitigate the impact of low-quality images and rectify inaccurately annotated labels. By leveraging this methodology, we achieve significant improvements in gaze estimation performance within the dataset.

In contrast, our approach uniquely addresses low-quality samples and inaccurately annotated labels in gaze estimation datasets. We propose a novel method to assess label consistency, estimating uncertainty in both image and label space. This tackles low-quality images and rectifies inaccurate labels, resulting in significant performance improvements.

\section{Method}

To mitigate the adverse impact of data uncertainty in both image space and label space of gaze estimation, we propose a novel Suppressing Uncertainty in Gaze Estimation (SUGE) method. 
In this section, we will begin by giving an overview of SUGE and subsequently introduce its three modules and co-training strategy in detail.

\subsection{Overview of SUGE}

\begin{figure*}
\centering
\includegraphics[width=17.6cm]{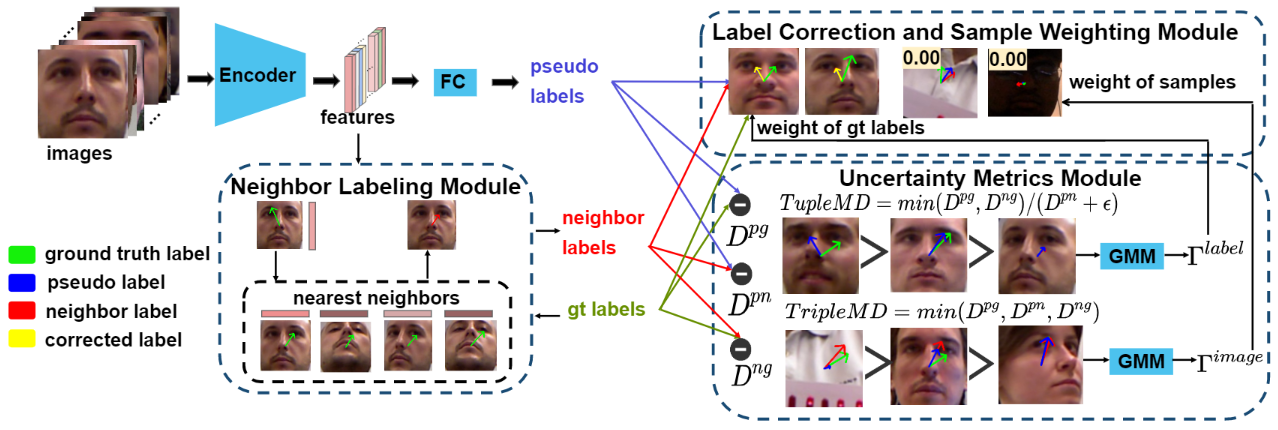}
\caption{The pipeline of SUGE method. Initially, input images undergo feature extraction through the encoder, and a fully connected layer generates pseudo labels. The neighboring labeling module then employs a nearest-neighbor algorithm to find feature neighbors for each image and calculates the neighboring label by weighted averaging the ground truth labels from its neighbors. Next, the Uncertainty Metrics module comes into play, computing Tuple Minimum Discrepancy and Triple Minimum Discrepancy by measuring the consistency among pseudo labels, ground truth labels, and neighboring labels. These uncertainties metrics are further input into a Gaussian Mixture Model, which yields two confidence scores: label confidence and image confidence. In the Label Correction and Sample Weighting module, label confidence is employed to perform weighted calculations on the ground truth labels, pseudo labels, and neighboring labels, resulting in corrected labels. Additionally, the sample weight is determined based on the image confidence and can further be used to guide the training process.}
\label{fig4.overview.png}
\end{figure*}

In order to capture the data uncertainty both in image and label space and further guide the following training process, we develop a novel triplet-label consistency for uncertainty estimation. The intuition behind our design is based on two foundations: 
% 1) for a given sample, the discrepancy between the ground-truth label and the predicted label can reflect the difficulty of fitting the current sample by the model; 2) the discrepancy between neighboring labels of a given sample and its two other types of labels can reflect the local similarity relationship between image features and their corresponding labels. 
1) The discrepancy between ground truth labels and pseudo labels for a given sample indicates its training difficulty and modeling complexity; 2) The discrepancy between neighboring labels and the other two types of labels reflects local relationship and smoothness between image features and their corresponding labels.
% Thus, observing the above triplet-label consistency can help us purify the quality of samples and the corresponding labels, mitigating overfitting risks and enhancing generalization.
By %upholding 
modeling the described triplet-label consistency, we can purify the quality of samples and their corresponding labels, which effectively reduces overfitting risks and enhances overall generalization.

To realize our idea, as depicted in the Fig.~\ref{fig4.overview.png}, our proposed SUGE is composed of three main modules: the neighboring labeling module, the uncertainty estimation module, as well as the label correction and sample weighting module. Specifically, for each input image, the encoder first extracts features, and the neighboring labeling module calculates the neighboring labels for the sample based on the ground truth labels of its neighbors in feature space. The uncertainty estimation module designs the uncertainty metrics of label and image by considering the consistency among neighboring labels, pseudo labels, and ground truth labels. Two uncertainty metrics are then input into the Gaussian Mixture Model (GMM)~\cite{permuter2006study} to calculate confidence scores. Finally, the label correction and sample weighting module utilizes the confidence scores to perform label correction and sample weighting for training.

Furthermore, to avoid overconfidence and cumulative errors during self-training of a single network on its own generated image and label confidence, we adopt the approach from the paper~\cite{friend1993co}~\cite{li2020dividemix} and train two networks simultaneously. Both networks contribute to generating corrected labels and sample weights for each other. The final algorithm is presented in Algorithm \ref{alg:algorithm}.

\begin{algorithm}[h!]
\caption{SUGE}
\label{alg:algorithm}
\textbf{Input}: two encoder and fully connected parameters $(\mathcal{E}^{(1)}, f^{(1)})$ and $(\mathcal{E}^{(2)}, f^{(2)})$, training data $(X,Y)$\\
\textbf{Param.}: small constant for denominator $\epsilon$, number of neighbors $K$, confidence threshold $\tau$
\begin{algorithmic}[1]
\STATE $(\mathcal{E}^{(1)}, f^{(1)}) \gets \text{WarmUp}(X,Y, (\mathcal{E}^{(1)}, f^{(1)}))$
\STATE $(\mathcal{E}^{(2)}, f^{(2)}) \gets \text{WarmUp}(X,Y, (\mathcal{E}^{(2)}, f^{(2)}))$
\WHILE{$e < \text{MaxEpoch}$}
\FOR{$k=1,2$} 
\STATE $\hat Y^{p} \gets f^{(k)}(\mathcal{E}^{(k)}(X))$
\STATE {\large// Neighbor labeling module}
\STATE $X_{ij} \gets \text{K-NN}(\mathcal{E}^{(k)}(X_i), K)$, for \(i = 1, \ldots, N\)
\STATE Get the reconstruction weight $A$ by Equation \ref{eq:LLR}
\STATE $\hat{Y}^{n}_{i} \gets \sum_{j=1}^{K} Y_{ij} * A_{ij}$, for \(i = 1, \ldots, N\)
\STATE {\large// Uncertainty estimation module}
\STATE $D^{pg}, D^{pn}, D^{ng} \gets \text{AngularDis}(Y, \hat Y^{n}, \hat Y^{p})$
\STATE $TupleMD \gets \frac{\min(D^{pg}, D^{ng})}{D^{pn} + \epsilon}$
\STATE $TripleMD \gets \min(D^{pg}, D^{pn}, D^{ng})$
\STATE $\Gamma^{\text{label}} \gets \text{GMM}(TupleMD)$
\STATE $\Gamma^{\text{image}} \gets \text{GMM}(TripleMD)$
\STATE {\large// Label correction and sample weighting}
\STATE $\Gamma^{\text{label}} \gets \text{Truncate}(\Gamma^{\text{label}}, \tau, 0)$
\STATE Calculate $\hat{Y}^{(k)}$ using $\Gamma^{\text{label}}$ by Equation \ref{eq:refine}
\STATE $\hat{\Gamma}^{(k)} \gets \Gamma^{\text{label}}$
\STATE $\Gamma^{\text{image}} \gets \text{Truncate}(\Gamma^{\text{image}}, \tau, 0)$
\STATE $\hat {W}^{(k)} \gets \Gamma^{\text{image}}$
\ENDFOR
\STATE If \(\hat {\Gamma}^{(1)}_i = 0\), set \(\hat{Y}^{(1)}_i =  \frac{\hat{Y}^{(1)}_i + \hat{Y}^{(2)}_i}{2}\), for \(i = 1, \ldots, N\)
\STATE If \(\hat {\Gamma}^{(2)}_i = 0\), set \(\hat{Y}^{(2)}_i =  \frac{\hat{Y}^{(1)}_i + \hat{Y}^{(2)}_i}{2}\), for \(i = 1, \ldots, N\)
\STATE $(\mathcal{E}^{(1)}, f^{(1)}) \gets \text{Train}(X, \hat{Y}^{(2)}, (\mathcal{E}^{(1)}, f^{(1)}), \hat W^{(2)})$
\STATE $(\mathcal{E}^{(2)}, f^{(2)}) \gets \text{Train}(X, \hat{Y}^{(1)}, (\mathcal{E}^{(2)}, f^{(2)}), \hat W^{(1)})$
\ENDWHILE
\end{algorithmic}
\end{algorithm}

\subsection{Neighboring Labeling Module}

In this module, we aim to capture the local similarity relationship between image features and their corresponding labels by computing neighboring labels. Formally, given a training dataset $\mathcal{D}=\{X,Y\}=\{(X_1,Y_1),...,(X_N,Y_N)\}$, where $X_i$ is $i^{th}$ input image and $Y_i=(Y_{i,\psi},Y_{i,\theta})^T$ is the ground truth 2D gaze angle vector (yaw and pitch), we use a $K$ nearest-neighbor (K-NN) algorithm to generate the $j^{th}$ neighbor of $X_i$ denoted by $X_{ij}, j=1...,K$ where the neighboring samples are identified based on the distance between samples in the feature space $\mathcal{E}(X)$, and are restricted to the same person ID as sample $X_i$.

Next, we find the optimal reconstruction weight $A_i$ for each sample $X_i$ with respect to its neighboring samples $X_{ij}, j=1...,K$. This is achieved by solving the following optimization problem:
\begin{equation}
\begin{aligned}
& \underset{A_i}{\text{minimize}}\, \| \mathcal{E}(X_i) - \sum_{j=1}^{K} (A_{ij} \mathcal{E}(X_{ij}))\|^2 + \lambda \|A_{i}\|^2, \\
& \text{subject to:} \quad \sum_{j=1}^{K} A_{ij} = 1.
\end{aligned}
\end{equation}

The closed-form solution for the optimal reconstruction weights is given by:
\begin{equation} \label{eq:LLR}
A_i = \frac{(S_i + \lambda I)^{-1} \mathbf{1}_K}{\mathbf{1}_K^T (S_i + \lambda I)^{-1} \mathbf{1}_K},
\end{equation}
where
%\begin{equation}
%\Delta(X_{i}) = \begin{bmatrix}
%\mathcal{E}(X_{i1})-\mathcal{E}(X_{i}) \\
%\mathcal{E}(X_{i2})-\mathcal{E}(X_{i}) \\
%\vdots \\
%\mathcal{E}(X_{iK})-\mathcal{E}(X_{i})
%\end{bmatrix},
%\end{equation}
\begin{equation*}
\Delta(X_{i}) = \begin{bmatrix}
\mathcal{E}(X_{i})-\mathcal{E}(X_{i1}), & \cdots & ,\mathcal{E}(X_{i})-\mathcal{E}(X_{iK})
\end{bmatrix},
\end{equation*}
\begin{equation}
S_i = \Delta(X_i) \Delta(X_i)^T,
\end{equation}
$I$ denotes the identity matrix, and $\mathbf{1}_K$ is a column vector containing $K$ elements with all elements equal to 1. 

Finally, we compute the new neighboring labels $\hat{Y^{N}_{i}}$ by element-wise multiplying the original labels $Y_{ij}$ with the optimal reconstruction weights $A_{ij}$:

\begin{equation}
\hat{Y}^{n}_{i} = \sum_{j=1}^{K} Y_{ij} * A_{ij}.
\end{equation}

Through above process, we obtain a new set of labels $\hat{Y}^n=\{\hat{Y}^n_{1},...,\hat{Y}^n_{N}\}$ called neighboring labels that capture the local similarity and smoothness between images and labels in the feature space.

\subsection{Uncertainty Estimation Module}

This module aims to assess the degree of uncertainty caused by low-quality labels and images by designing suitable metrics based on the consistency among the predicted pseudo label \(\hat{Y}_i^p=f(\mathcal{E}(X_i))\), the neighboring label \(\hat{Y}_i^n\), and the ground truth label \(Y_i\) for a given sample $X_i$. We focus on the angular differences among these label types, denoting them as \(D_i^{pg}\), \(D_i^{pn}\), and \(D_i^{ng}\) for measuring the triplet-label consistency. Here taking \(D_i^{pg}\) as an example, \(D_i^{pn}\) and \(D_i^{ng}\) can be given following the similar calculation.

Firstly, we convert the labels from polar coordinate system to 3D Cartesian coordinate system using the gazeto3D function:
\begin{equation}
{\text{gazeto3d}(\hat{Y}_i^p) = \begin{bmatrix}
-\cos(\hat{Y}_{i,\theta}^p) \cdot \sin(\hat{Y}_{i,\psi}^p) \\
-\sin(\hat{Y}_{i,\theta}^p) \\
-\cos(\hat{Y}_{i,\theta}^p) \cdot \cos(\hat{Y}_{i,\psi}^p)
\end{bmatrix}},
\end{equation}

\begin{equation}
{\text{gazeto3d}(Y_i) = \begin{bmatrix}
-\cos(Y_{i,\theta}) \cdot \sin(Y_{i,\psi}) \\
-\sin(Y_{i,\theta}) \\
-\cos(Y_{i,\theta}) \cdot \cos(Y_{i,\psi})
\end{bmatrix}}.
\end{equation}

Next, we compute the angular distance between the two 3D coordinates using the angular function:
\begin{equation} \label{eq:angular_distance}
D_i^{pg} = \arccos\left(\frac{\text{gazeto3d}(\hat Y^p_i) \cdot \text{gazeto3d}(Y_i)}{\lVert \text{gazeto3d}(\hat Y^p_i)  \rVert \cdot \lVert \text{gazeto3d}(Y_i) \rVert}\right) \cdot \frac{180}{\pi},
\end{equation}
where \(\lVert \cdot \rVert\) denotes the Euclidean norm.

% Building upon the above differences among three labels, we believe that reliable ground truth labels should be at least close to either the pseudo labels or the neighboring labels. Therefore, we introduce a metric named as Tuple Minimum Discrepancy (TupleMD) to measure the degree of uncertainty related to the label quality. The TupleMD is positively related to the minimum difference between the ground truth labels and the other two types of labels. Additionally, it is inversely correlated with their discrepancy, reflecting the similarity and accuracy of the information provided by pseudo labels and neighboring labels. The formula for \(TupleMD\) is as follows:
% \begin{equation}
% TupleMD_i = \frac{\min(D^{pg}_i, D^{ng}_i)}{D^{pn}_i + \epsilon},
% \end{equation}
% where \(\epsilon\) is a small constant used to prevent excessively large label uncertainty when \(D^{pn}_i\) approaches zero.

To measure the degree of uncertainty related to the label quality, we propose \textbf{TupleMD} as follows:
\begin{equation}
TupleMD_i = \frac{\min(D^{pg}_i, D^{ng}_i)}{D^{pn}_i + \epsilon}.
\end{equation}
The numerator of \textbf{TupleMD} is the minimum difference between the GT label and the other two labels (the pseudo label and neighboring label),which means that a GT label is considered unreliable when it exhibits substantial discrepancies with both two other labels. The denominator of \textbf{TupleMD} is a scale factor defined by the distance between the pseudo label and the neighboring label, reflecting the uncertainty degree they contribute to the numerator. In addition to $D^{pn}_{i}$, the denominator also includes a constant term $\epsilon$, to ensure that uncertainty is not erroneously considered high when both the numerator and denominator are very small.

% Furthermore, we define another metric, named Triple Minimum Discrepancy (TripleMD), to represent the degree of uncertainty related to image quality. TripleMD should be positively related to the minimum difference among these three types of labels, indicating those samples with large TripleMD might be disturbed by noise. We formulate the \(TripleMD\) as follows:
% \begin{equation}
% TripleMD_i = \min(D^{pg}_i, D^{pn}_i, D^{ng}_i).
% \end{equation}

Furthermore, to represent the degree of uncertainty related to image quality, we propose \textbf{TripleMD} as follows:
\begin{equation}
TripleMD_i = \min(D^{pg}_i, D^{pn}_i, D^{ng}_i).
\end{equation}
The motivation behind this design is that the proposed three labels can be viewed as the representations provided by three `experts', each offering their findings for the given image. GT labels and pseudo labels reflect the difficulties in data collection and model training respectively, and neighboring labels represent the relationship between similar images. 
% While gaze is inherently deterministic, the uncertainty in observations is introduced during the data collection process, resulting in  low-quality images. 
Therefore, if any two findings exhibit significant discrepancies for a particular image, we treat it as a low-quality image.%Therefore, we employ the metric $\min(D_{i}^{pg}, D_{i}^{ng}, D_{i}^{pn})$ to assess this level of uncertainty.}

Finally, to determine the probabilities of the above two uncertainty metrics, we adopt the bimodal Gaussian Mixture Model (GMM) to partition the two uncertainty metrics \(TupleMD\) and \(TripleMD\) into reliable and unreliable components. For each sample \(X_i\), we define the label confidences \(\Gamma^{label}_i\) as the posterior probability \(p(g|TupleMD_i)\) and image confidence \(\Gamma^{image}_i\) as the posterior probability \(p(g|TripleMD)\), where \(g\) represents the Gaussian component with the smaller mean, indicating the likelihood of the sample being considered reliable. This allows us to quantify the uncertainty and assess the quality of the labels and images in the dataset.

\subsection{Label Correction and Sample Weighting Module}
In this module, we generate corrected labels and sample weights based on the confidences of labels and images for better training. We observe that the majority of label and image confidences are close to 1, while only a small number of confidence scores are low, indicating severe issues. To address this, for each sample $X_i$ with confidence $\Gamma_i$, we set a threshold and truncate low confidences below the threshold to zero:
\begin{equation}
\Gamma_i = \begin{cases}
\Gamma_i & \text{if} \ \Gamma_i > \tau \\
0 & \text{otherwise}
\end{cases},
\end{equation}
where \(\Gamma_i\) represents the label or image confidence of $X_i$, and \(\tau\) is the threshold value.

% \textbf{Label correction.} Using the updated label confidence \(\Gamma_i^{label}\), we calculate a new rectified label by combining the ground truth label \(Y_i\) with the pseudo label \(\hat Y_i^p\) and neighboring labels \(\hat Y^n_i\). To improve the label accuracy, data augmentation is performed on the original images \(X_i\) to obtain the augmented pseudo label \(\hat Y_i^{pa}\):
% \begin{equation}
% \hat{Y}_i^{pa} = f(\mathcal {E}(\text{Augmentation}(X_i))),
% \end{equation}
% where \(\mathcal {E}\) represents the encoder and \(f\) represents the fully connected layer.

\textbf{Label correction.} With the updated label confidence \(\Gamma_i^{label}\), we can correct the ground truth label \(Y_i\) by  pseudo label \(\hat Y_i^p\) and neighboring label \(\hat Y^n_i\). To further enhance the accuracy of the corrected label, we employ the widely used gaze estimation augmentation technique of horizontal flipping on the original images \(X_i\), maintaining unchanged eye gaze pitch while inverting the yaw component. This yields the augmented pseudo label \(\hat Y_i^{pa}\):
\begin{equation}
\hat{Y}_i^{fpa} = f(\mathcal {E}(\text{HorizontalFlip}(X_i))),
\end{equation}
\begin{equation}
\hat{Y}_{i}^{pa}= (-\hat{Y}_{i,\psi}^{fpa}, \hat{Y}_{i,\theta}^{fpa})^T,
\end{equation}
where \(\mathcal {E}\) represents the encoder and \(f\) represents the fully connected layer.

Next, we use the neighboring sample \(X_{ij}(j=1...K)\) and the corresponding reconstruction weight \(A_{ij}\)  of sample \(X_i\) obtained from the neighboring labeling module to calculate pseudo neighboring labels \(\hat {Y}^{np}_i\) and pseudo augmented neighboring labels \(\hat {Y}^{npa}_i\) based on \(\hat {Y}^p_i\) and \(\hat{Y}^{pa}_i\). The formulations are as follows:
\begin{equation}
\hat{Y}^{np}_{i} = \sum_{j=1}^{K} \hat Y^{p}_{ij} * A_{ij},
\hat{Y}^{npa}_{i} = \sum_{j=1}^{K} \hat Y_{ij}^{pa} * A_{ij}.
\end{equation}

Finally, the corrected labels are computed as a combination of the ground truth labels and the generated pseudo labels:
\begin{equation}
\hat{Y}_i = \Gamma^{\text{label}}_i \cdot Y_i + (1 - \Gamma^{\text{label}}_i) \cdot \frac{1}{5} (\hat Y^n_i + \hat Y^{p}_i + \hat Y^{pa}_i + \hat Y^{np}_i + \hat Y^{npa}_i).
\label{eq:refine}
\end{equation}

{\textbf{Sample weighting.} Additionally, to reduce the effect of low-quality samples, we further exploit the image confidence as a weight $\hat{W}_i = \Gamma^{\text{image}}_i$ to guide the training process. The overall loss objective is as follows: }
\begin{equation}
%\hat{W}_i = \Gamma^{\text{image}}_i.
{\mathcal{L} = \sum_{i} \hat{W}_i \cdot {\lVert \hat{Y}_i-f(\mathcal{E}(X_i) \lVert}_1}.
\end{equation}

\subsection{Co-training Strategy}
To further alleviate the data uncertainty both in image and label space, we follow ~\cite{friend1993co}~\cite{li2020dividemix} to introduce co-training strategy for maintaining two networks simultaneously, where they exchange the corrected labels \(\hat{Y}\) and sample weights \(\hat{W}\) in each iteration to prevent excessive confidence in their self-evaluation. Furthermore, for labels with truncated confidence equaling to zero, we additionally correct them by averaging the corrected labels generated by both networks.

\section{Experiment}
\subsection{Dataset}
The experiments utilize four widely used gaze estimation datasets: EyeDiap~\cite{funes2014eyediap}, MPIIFaceGaze~\cite{zhang2017mpiigaze}, Gaze360~\cite{kellnhofer2019gaze360}, and ETH-XGaze~\cite{zhang2020eth} (solely utilized for pretraining the GazeTR model). %To ensure fair comparisons, 
For a fair comparison, the data partitioning and preprocessing techniques for these datasets are maintained consistently with prior studies, as outlined by Cheng et al.~\cite{cheng2021appearance}.

% The \textit{EyeDiap} dataset consists of 16,000 images collected from 14 users in a controlled laboratory environment and undergoes a four-fold cross-validation. The \textit{MPIIFaceGaze} dataset includes 16,000 images taken from 14 users under complex lighting conditions and employs a leave-one-out training strategy. The \textit{Gaze360} dataset encompasses 84,000 training images from 54 subjects and 16,000 testing images from 15 subjects, showcasing a wide range of gaze directions. The \textit{ETH-XGaze} dataset comprises 1,083,492 high-resolution images collected from 110 users, exclusively employed for pretraining the GazeTR model in our experiment. The \textit{ETH-XGaze} dataset solely utilized for pretraining the GazeTR model.

% The \textit{EyeDiap} dataset consists of 16,534 images collected from 14 users and undergoes a four-fold cross-validation. The \textit{MPIIFaceGaze} dataset includes 45,000 images taken from 15 users under complex lighting conditions and employs a leave-one-out training strategy. The \textit{Gaze360} dataset encompasses 84,000 training images from 54 subjects and 16,000 testing images from 15 subjects, showcasing a wide range of gaze directions. The \textit{ETH-XGaze} dataset comprises 1,083,492 high-resolution images collected from 110 users, solely utilized for pretraining the GazeTR model.

\subsection{Implementation Details}

Since our approach primarily focuses on mitigating data uncertainty through label correction and weighting, regardless of specific network architectures, we directly adopt two representative state-of-the-art (SOTA) methods namely Gaze360~\cite{kellnhofer2019gaze360} and GazeTR~\cite{cheng2022gaze} implemented by Cheng et al.~\cite{cheng2021appearance} as baselines in our subsequent experiments. We utilize the same network architecture and corresponding parameter settings as these methods. Only for addressing the challenge of high-dimensional feature clustering, we reduce the feature dimension to 16 at the final layer of the encoder.

Regarding our method, we set \(\epsilon\) to 1 to prevent excessive \(TupleMD\). Additionally, for the $K$ nearest neighbor algorithm, we choose the KD tree with $K=4$. 
%we choose the KD tree, with the default number of clusters \(K\) set to 4. 
Moreover, we set the warm-up epochs to 10, and the thresholds \(\tau\) for truncating label and image confidences are both set to 0.5.

\subsection{Comparison with SOTA Gaze Estimation Methods}

We compare our method with state-of-the-art gaze estimation methods, including classical CNN-based methods pretrained on ImageNet, such as FullFace~\cite{zhang2017s}, CA-Net~\cite{cheng2020coarse} and Gaze360~\cite{kellnhofer2019gaze360}, as well as Transformer-based methods pretrained on ETH-XGaze, such as CADSE~\cite{o2022self} and GazeTR~\cite{cheng2022gaze}. Our method focuses on purifying the data and labels, which can be easily combined with other models, so we re-implement the leading open-source methods from both categories, namely Gaze360$^\dagger$ and GazeTR$^\dagger$, and verify the effectiveness of our proposed method by assessing their performance with and without our SUGE approach on the EyeDiap, MPIIFaceGaze, and Gaze360 datasets.

% As shown in Table \ref{table:Gaze_estimation_performance}, compared with CNN-based Gaze360, our approach effectively reduces angle errors by 0.1$^{\circ}$, 0.26$^{\circ}$, and 0.41$^{\circ}$ on MPII, Gaze360 and EyeDiap datasets. Compared with Transformer-based GazeTR, the angle errors are also largely reduced by ***$^{\circ}$, 0.10$^{\circ}$, and 0.30$^{\circ}$ on MPII, Gaze360 and EyeDiap datasets. These results significantly suggest that our approach can  
% improve the quality of datasets, thereby ultimately achieving the most up-to-date SOTA results in gaze estimation task.

As shown in Table \ref{table:Gaze_estimation_performance}, compared with CNN-based Gaze360$^\dagger$, our approach reduces angle errors by 0.1$^{\circ}$, 0.26$^{\circ}$, and 0.41$^{\circ}$ on MPII, Gaze360 and EyeDiap datasets. Compared with Transformer-based GazeTR$^\dagger$, the angle errors remain consistent on MPII dataset, while they are notably reduced by 0.10$^{\circ}$ and 0.30$^{\circ}$ on Gaze360 and EyeDiap datasets. These results significantly suggest that our approach can  
improve the quality of datasets, thereby ultimately achieving the most up-to-date SOTA results in gaze estimation task.

\begin{table}[t]
  \tabcolsep=1.0mm
  \renewcommand\arraystretch{0.9}
  \centering
  \begin{tabular}{@{}lcccc@{}}
    \toprule
    \multirow{2}{*}{Method} & \multirow{2}{*}{Backbone} & \multicolumn{3}{c}{Dataset} \\
    \cmidrule(lr){3-5}
     & & MPII & Gaze360 & EyeDiap \\
    \midrule
    Full Face & C & 4.93 & 14.99 & 6.53 \\
    CA-Net & C & 4.27 & 11.20 & 5.27 \\
    Gaze360 & C & 4.06 & 11.04 & 5.36 \\
    CADSE  & T & 4.04 & 10.70 & 5.25 \\
    GazeTR & T & 4.00 & 10.62 & 5.17 \\
    \midrule
    Gaze360$^{\dagger}$ & C & 4.17 & 10.78 & 5.46 \\
    GazeTR$^{\dagger}$ & T & 4.00 & 10.61 & 5.34 \\
    \midrule
    SUGE (Gaze360) & C & 4.07 & 10.52 & 5.05 \\
    SUGE (GazeTR) & T & 4.01 & 10.51 & 5.04 \\
    \bottomrule
  \end{tabular}
  \caption{Comparison of gaze estimation performance in terms of angle error ($^\circ$) on three datasets.  $^{\dagger}$represents our re-implemented results. %B represents Backbone (CNN: C, Transformer: T), and P represents Pretrain (Imagenet: I, ETH-XGaze: E).
  C and T represent CNN (pretrained by Imagenet) and Transformer (pretrained by ETH-XGaze) backbone respectively.}
  \label{table:Gaze_estimation_performance}
\end{table}

\subsection{The Effectiveness of Our Uncertainty Estimation and Label Correction Paradigm}%Comparison with Famous Noisy Label Methods}
The issue of low-quality labels in gaze estimation is analogous to the classic problem of noisy label learning. To demonstrate the superiority of our design, we adopt the popular noise label learning methods on the Gaze360 baseline model and compare them with our proposed strategy on Gaze360 and EyeDiap datasets. It should be noted that most latest methods~\cite{li2023disc}~\cite{wei2023fine} are basically designed for classification problems and can not be suitable for our gaze regression task.
Therefore, we carefully choose the most notable noise learning strategies: Co-Teaching~\cite{han2018co} and DivideMix~\cite{li2020dividemix} for verification. These two strategies can be readily adapted for regression tasks since they directly utilize the loss as the sample selection criterion. The comparisons are conducted using the Gaze360 and EyeDiap datasets. Remarkably, DivideMix still maintains its preeminence as a leading method in the most real-world benchmarks of learning from noisy labels, like Clothing 1M.
% Further details on the implementation can be found in the supplementary material.

\begin{table}
\centering
\begin{tabular}{lcc}
\toprule
\multirow{2}{*}{Method} & \multicolumn{2}{c}{Dataset} \\
\cmidrule(lr){2-3}
 & Gaze360 & EyeDiap \\
\midrule
Baseline & 10.78 & 5.46 \\
CoTeaching & 10.59 & 5.18 \\
DivideMix & 10.66 & 5.16 \\
SUGE & 10.52 & 5.05 \\
\bottomrule
\end{tabular}
\caption{Comparison in terms of angle error ($^\circ$) on two datasets with noise label learning methods.}
\label{table:Noise_label_comparison}
\end{table}

The comparisons with other label noisy methods are summarized in Table~\ref{table:Noise_label_comparison}. We can see that our method outperforms two excellent noise label methods in the gaze estimation task, demonstrating that our proposed uncertainty estimation can successfully access the quality of data and label and further effectively guide the following training process. Moreover, all noise label learning strategies, including our SUGE, surpass the baseline model, highlighting the importance of addressing low-quality labels in the gaze estimation task.

\subsection{Ablation Study}
Our SUGE method benefits from its three core components. Firstly, it involves neighboring labels combined with pseudo labels and ground-truth labels to assess data uncertainty via triplet-label consistency. Secondly, it designs image and label confidences for label correction and sample weighting. Thirdly, it employs the co-training strategy for alleviating the negative effects of label noise. We conduct ablation experiments using the Gaze360 baseline model on the EyeDiap datasets to reveal the effects of these components.

\begin{table}
\centering
\tabcolsep=1mm
\begin{tabular}{lc}
\toprule
Method & Angle Error ($^\circ$) \\
\midrule
%neighbor average label
w/o neighboring labeling & 5.16 \\
w/o reconstruction weighting & 5.10 \\
\midrule
w/o sample weighting & 5.10 \\
w/o label correction & 5.17 \\
w/o sample weighting \& label correction  & 5.23 \\
\midrule
 w/o co-training & 5.14 \\
\midrule
w/o entire label compositions & 5.10\\
\midrule
SUGE & 5.05 \\
\bottomrule
\end{tabular}
\caption{Ablation studies %using Gaze360 Model 
on EyeDiap dataset.}
\label{table:ablation_study}
\end{table}

\subsubsection{Neighboring Labeling.} To validate the importance of introducing neighboring labels, we conduct two ablations. In the first ablation, we completely remove the neighboring labeling module, and only perform label correction based on the consistency between ground truth labels and pseudo labels, which is similar to the DivideMix~\cite{li2020dividemix} approach. In the second ablation, we only remove the reconstruction weighting calculation and directly calculate the average of neighbors' labels to obtain the neighboring labels.

In Table~\ref{table:ablation_study}, we can see that removing the neighboring labeling modules (1st row) leads to an obvious performance drop (0.11$^{\circ}$). Besides, the weighting reconstruction strategy (2nd row) can also bring a performance gain (0.05$^{\circ}$).

\subsubsection{Label correction and sample weighting.} To evaluate the individual contributions of the label correction and sample weighting modules, we separately remove sample weighting and label correction modules. Besides, we also exclude both label correction and sample weighting, which can be viewed as an ensemble of two networks.
Table~\ref{table:ablation_study} shows the results. It can be seen that removing sample weighting (3rd row) results in a performance drop (0.05$^{\circ}$), and removing label correction (4th row) brings a larger drop (0.12$^{\circ}$). Obviously, combining two strategies for training (5th row) achieves the best performance gain (0.18$^{\circ}$), 
demonstrating the necessity of the label correction and sample weighting modules.

\subsubsection{Co-training strategy.} To mitigate accumulated errors, we adopt the co-training strategy where two networks feed data samples for each other during training. To validate the effect of co-training strategy, we directly employ the self-training strategy where two networks are trained separately. The results presented in the 6th row of Table~\ref{table:ablation_study} indicate that the angle error of self-training is higher than co-training (0.09$^{\circ}$), confirming the effectiveness of the co-training strategy.

\subsubsection{Label compositions for GT label correction} To evaluate the necessity of using the entire label composition in equation \ref{eq:refine}, we conduct an experiment using its subset, represented as $\hat{Y}_i = \Gamma^{\text{label}}_i \cdot Y_i + (1 - \Gamma^{\text{label}}_i) \cdot \frac{1}{2} (\hat Y^n_i + \hat Y^{p}_i)$. Table~\ref{table:ablation_study} illustrates that utilizing this subset (7th row) results in a higher angular 
error (0.05$^{\circ}$), confirming the importance of utilizing the entire label composition.
\subsection{Parameter Sensitivity Analysis}
Our SUGE method primarily consists of three hyperparameters: the number of neighbors \(K\), label confidence threshold \(\tau^{label}\), and image confidence threshold \(\tau^{image}\). We analyze the impact of hyperparameter settings in this experiment. As shown in the Table ~\ref{table:parameter_analysis_results}, we can observe that within a certain range, adjusting the hyperparameters has a relatively stable impact on the performance, and $K=4, \tau^{image}=0.5, 
\tau^{label}=0.5$ achieves the best performance.

\begin{table}
\centering
\begin{tabular}{ccc}
%\hline
\toprule
Parameter & Value & Angle Error ($^\circ$)\\
%\hline
\midrule
 & 2 & 5.08 \\
$K$  & 4 & 5.05 \\
 & 6 & 5.09 \\
%\hline
\midrule
 & 0.4 & 5.11 \\
 $\tau^{image}$ & 0.5 & 5.05 \\
 & 0.6 & 5.09 \\
%\hline
\midrule
 & 0.4 & 5.14 \\
$\tau^{label}$ & 0.5 & 5.05 \\
 & 0.6 & 5.13 \\
%\hline
\bottomrule
\end{tabular}
\caption{Parameter analysis %Results using Gaze360 Model 
on EyeDiap dataset.}
\label{table:parameter_analysis_results}
\end{table}
\vspace{-5pt}

\subsection{Visualization Results}
\begin{figure}
\centering
\includegraphics[width=8.3cm]{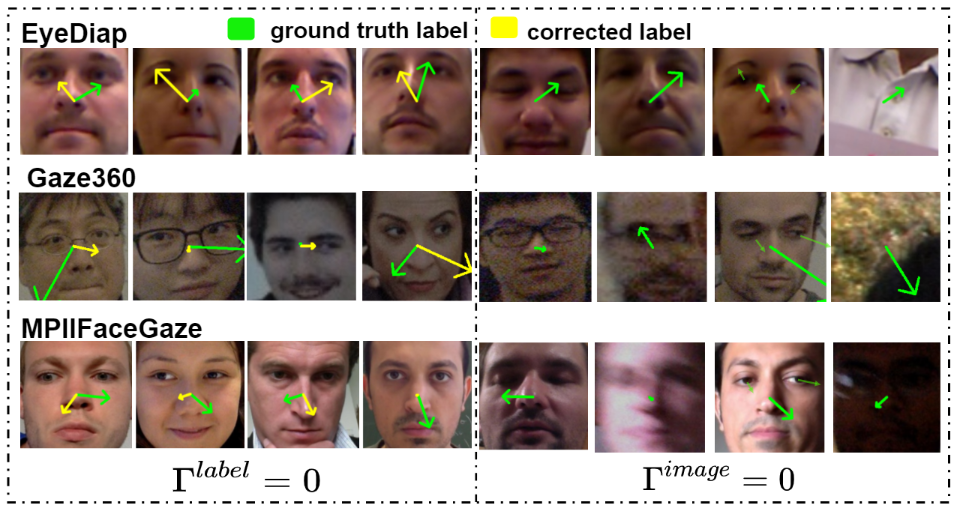}
\caption{Visualization of samples with label confidences set to 0 on the left figure and image confidences set to 0 on the right figure.  These results are sourced from the EyeDiap dataset (folds 0, 1, 3 as the training set), Gaze360 dataset, and MPIIFaceGaze dataset (users 1-14 as the training data), during the initial epoch after warm-up.}
\label{label:vis_png}
\end{figure}

In this subsection, we illustrate the effectiveness of our approach in reducing uncertainty of gaze estimation by visualizing some samples, where either their label confidences are set to 0 during training, indicating that these samples are initially annotated as wrong labels, and will be rectified with correct gaze directions, or their image confidences are set to 0, showing that these are low-quality samples and will be discarded during training. As shown in the left of Fig.~\ref{label:vis_png}, our method accurately identifies and corrects wrong annotated labels. Moreover, as depicted in the right of Fig. \ref{label:vis_png}, our approach demonstrates a strong capability to identify low-quality images caused by various adverse factors. %More  visualization results are provided in  supplementary material. 

%For more comprehensive visualization results, please refer to the supplementary material.
% \begin{figure}
% \centering
% \includegraphics[width=8cm]{Fig6.noisy_label.png}
% \caption{Visualization of ground truth labels (green) and corrected labels (yellow) for samples with label confidence set to 0. The results above are obtained from the EyeDiap dataset using folds 0, 1, and 3 as the training set in the first epoch after warm-up. Similarly, the results above are derived from the Gaze360 training dataset after the first epoch following warm-up.}
% \label{label:noisy_label}
% \end{figure}

% \begin{figure}
% \centering
% \includegraphics[width=8cm]{Fig5.low_quailty.png}
% \caption{The visualization of low-quality images, caused by occlusions, blurriness, inconsistent gaze directions between the eyes, and non-face images. These results are obtained from the EyeDiap dataset using folds 0, 1, and 3 as the training set in the first epoch following warm-up, as well as from the Gaze360 training dataset after the same epoch.}
% \label{label:noisyimage}
% \end{figure}

\vspace{-10pt}

\section{Conclusion}
In this study, we firstly discover the data uncertainty caused by data collection and annotation process in gaze estimation datasets, which is ignored by existing works. To reduce the negative effects of low-quality images and incorrect labels, we propose a novel approach named SUGE, which adopts the triplet-label consistency to estimate the uncertainty and subsequently utilizes it to guide the training process. %The experiments conducted on benchmarks demonstrate that our method significantly improves the performance of gaze estimation. 
% refine labels and assign weights to individual samples. 
The comprehensive  experiments conducted on popular benchmarks demonstrate that our method significantly improves the performance of gaze estimation. 

\bigskip

\section{Acknowledgements}
This work is supported by National Natural Science Foundation of China (62271042, 62376021, 62302032, 62106017), Beijing Natural Science Foundation (M22022, L211015, 4232032), and Hebei Natural Science Foundation (F2022105018).

\bibliography{aaai24}

\end{document}